\documentclass[10pt,twocolumn,letterpaper]{article}

\usepackage[pagenumbers]{iccv} % To force page numbers, e.g. for an arXiv version

\newcommand{\methodnameacr}{I2R\xspace}
\newcommand{\methodName}{Invert2Restore\xspace}

\definecolor{iccvblue}{rgb}{0.21,0.49,0.74}
\usepackage[pagebackref,breaklinks,colorlinks,allcolors=iccvblue]{hyperref}

\usepackage{algorithm}
\usepackage{algorithmic}
\usepackage{tabularx}
\usepackage{float}

\usepackage{bbm}
\usepackage{dsfont}

\newtheorem{proposition}{Proposition}

\author{%
Hamadi Chihaoui \quad  Paolo Favaro \\
Computer Vision Group, University of Bern, Switzerland\\
\texttt{\{hamadi.chihaoui,paolo.favaro\}@unibe.ch}\\
}

\title{Invert2Restore: Zero-Shot Degradation-Blind Image Restoration}
\begin{document}
\maketitle

\begin{abstract}
%{\color{red}{Slight change to account for the fact that we can also handle the partially-blind case}}
Two of the main challenges of image restoration in real-world scenarios are the accurate characterization of an image prior and the precise modeling of the image degradation operator. Pre-trained diffusion models have been very successfully used as image priors in zero-shot image restoration methods. However, how to best handle the degradation operator is still an open problem. 
%Most image restoration methods are either non-blind (the degradation model is completely known) or partially blind (the parametric form of the degradation model is known, but not its parameters). 
In real-world data, methods that rely on specific parametric assumptions about the degradation model often face limitations in their applicability. To address this, we introduce \methodName, a zero-shot, training-free method that operates in both fully blind and partially blind settings -- requiring no prior knowledge of the degradation model or only partial knowledge of its parametric form without known parameters. Despite this, \methodName achieves high-fidelity results and generalizes well across various types of image degradation. It leverages a pre-trained diffusion model as a deterministic mapping between normal samples and undistorted image samples.
%In real-world data, methods that make any parametric assumption about the degradation model limit their applicability. Thus, we introduce \methodName, a zero-shot, training-free and fully blind method that requires no prior knowledge of the degradation model, yet delivers high-fidelity results and generalizes well across various types of image degradation. \methodName uses a pre-trained diffusion model as a deterministic mapping between normal samples and undistorted image samples. 
The key insight is that the input noise mapped by a diffusion model to a degraded image lies in a low-probability density region of the standard normal distribution. Thus, we can restore the degraded image by carefully guiding its input noise toward a higher-density region. We experimentally validate \methodName across several image restoration tasks, demonstrating that it achieves state-of-the-art performance in scenarios where the degradation operator is either unknown or partially known.
%Recently, many zero-shot image restoration methods based on pre-trained diffusion models have been proposed. However, most of these methods are either non-blind (assuming that the degradation model is known) or partially blind (assuming the parametric form of the model is known), which may limit their applicability in real-world degradation scenarios. In this paper, we introduce \methodName, a zero-shot, training-free and fully blind method that requires no prior knowledge of the degradation model, yet delivers high-fidelity results and generalizes well across various types of image degradation. \methodName works by embedding the degraded image back into the initial input noise space of pre-trained diffusion models. The key insight is that, in the case of a degraded image, the initial input noise originates from a low-density region of the standard normal distribution. We restore the image by carefully moving it toward a higher-density region. We experimentally validate \methodName across several image restoration tasks, demonstrating that it achieves state-of-the-art performance in scenarios where the degradation prior is unknown.
\end{abstract}

\section{Introduction}
%{\color{red}{Slight change to account for the fact that we can also handle the partially-blind case}}

Image restoration (IR) is a long-standing inverse problem in imaging, where the goal is to recover a so-called \emph{clean} image (\ie, an image without distortions) from a \emph{degraded} one \cite{Zhu_2023_CVPR,li2023diffusionmodelsimagerestoration}. 
Recently, deep learning techniques have achieved notable success in this area, with methods generally falling into two categories: training-based \cite{liang2021swinir, wang2021real} and training-free approaches \cite{kawar2022denoising, lugmayr2022repaint, chung2022improving, chung2022diffusion}.
%Supervised IR models are typically trained on large, paired datasets, assuming that both training and testing data share the same distribution. A key limitation is that these models tend to perform poorly when the test data differs from the training set. Moreover, whenever the degradation process changes, a new dataset must be collected and the model retrained, which can be resource-intensive.
%In contrast, unsupervised IR techniques utilize degradation models to restore clean images by solving optimization or sampling problems. For instance, methods like DDRM rely on a linear degradation model to sample from the posterior distribution. However, real-world degradation models can be far more complex and computationally difficult to handle, limiting the effectiveness of such methods. 
Training-based IR methods require a dataset and can be classified into supervised and unsupervised approaches. Traditional supervised methods often rely on large datasets of paired clean and degraded images, which are costly and time-consuming to collect. Unsupervised training-based methods relax this requirement by eliminating the need for labeled data. However, training-based methods tend to generalize poorly at test time on new data with degradations not matching those in the training set. Moreover, these approaches can be quite costly as the whole pipeline of collecting data and training a model must be repeated whenever the degradation process changes. 

% \begin{figure}[t]
% \centering\
%    \setkeys{Gin}{width=0.33\linewidth}
%     \captionsetup[subfigure]{skip=0.0ex,
%                              belowskip=0.0ex,
%                              labelformat=simple}
%                              \setlength{\tabcolsep}{1.0pt}
%     \renewcommand\thesubfigure{}
%     \small
%     \small

% \small
%   \begin{tabular}{ccc}
 
%  {\includegraphics{images/teaser1/oldphoto3.png}}& {\includegraphics{images/teaser1/dreamclean111.png}} &{\includegraphics{images/teaser1/3.png}}\\
%   \textrm{Old photo} & \textrm{DreamClean~\cite{xiaodreamclean}} & \textrm{\methodnameacr} \\
  
%    \end{tabular}
% \captionof{figure}{\label{fig:teaser} In real-world images, the degradation can be unknown, complex and not easy to model, such as in the case of the restoration of old photos. Compared to other methods, \methodnameacr generates sharper and more identity-preserving outputs. %\vspace{.5em}
% }
% \end{figure}

In contrast, training-free image restoration (IR) methods have garnered significant attention in recent years due to their ability to restore degraded images without relying on task-specific training data and their adaptability to target degraded images at test time. These approaches leverage internal data priors from a single degraded image or pre-trained models to perform restoration tasks. However, many of the recently proposed training-free approaches \cite{kawar2022denoising, lugmayr2022repaint, chung2022improving, chung2022diffusion} are non-blind, meaning that they assume full knowledge of the degradation model, including both its parametric form and specific parameters, a strong assumption that limits their applicability in real-world scenarios. Recently, some partially-blind training-free methods \cite{chung2023parallel, fei2023generative,chihaoui2024blind} have been proposed. These methods assume that the parametric form of the degradation is known and estimate the parameters directly from the given sample. However, in real-world cases, even the parametric form of the degradation process may be unknown and complex, as seen with weather-related degradations (\eg, rain and snow) or JPEG artifact removal.

\begin{figure*}[t]
\centering\
   \setkeys{Gin}{width=0.15\linewidth}
    \captionsetup[subfigure]{skip=0.0ex,
                             belowskip=0.0ex,
                             labelformat=simple}
                             \setlength{\tabcolsep}{1.pt}
    \renewcommand\thesubfigure{}
    \small

\small
  \begin{tabular}{cccccc}
  \textrm{JPEG artifacts} & \textrm{Raindrop}  &  \textrm{ Blur} & \textrm{Rain}& \textrm{$8\times$ SR}& \textrm{Old photo}  \\
 {\includegraphics{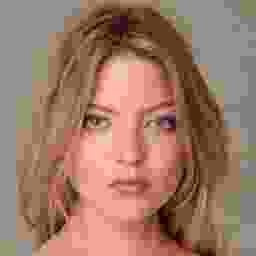}}&{\includegraphics{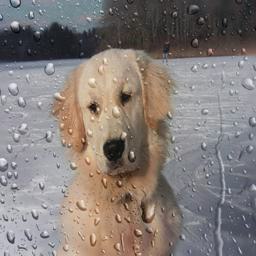}}& {\includegraphics{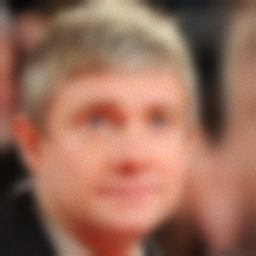}}&
 {\includegraphics{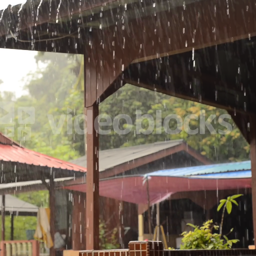}}&   
 {\includegraphics{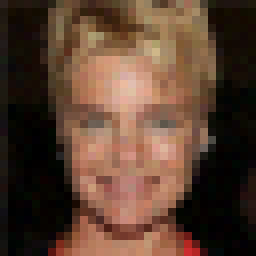}}&  {\includegraphics{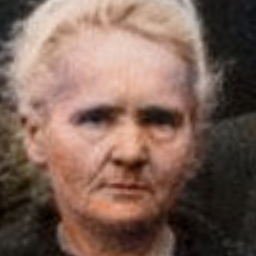}}\\
 {\includegraphics{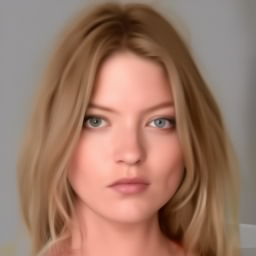}}& 
 {\includegraphics{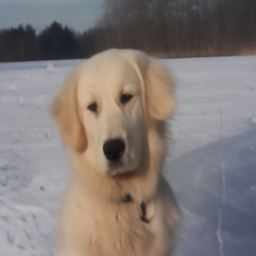}}& {\includegraphics{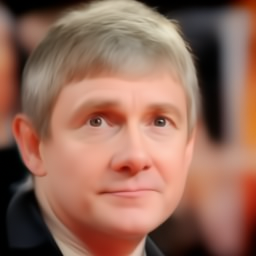}} & 
{\includegraphics{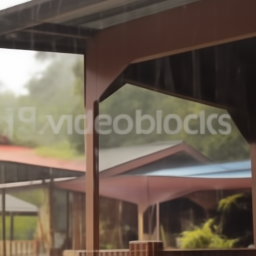}}& 
 {\includegraphics{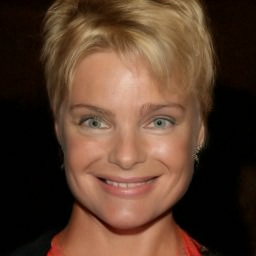}}&  {\includegraphics{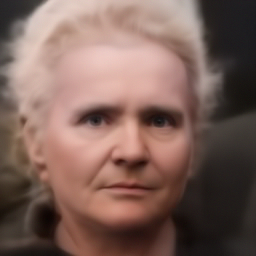}} \\
 {\includegraphics{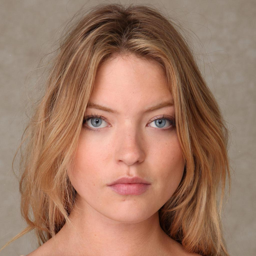}}&
 {\includegraphics{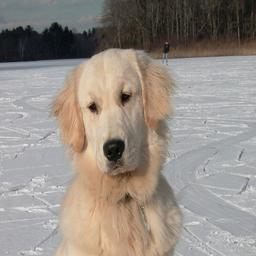}}&  {\includegraphics{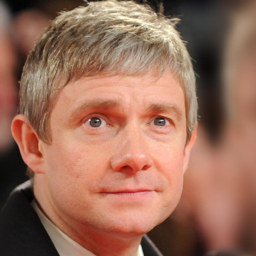}}& {\includegraphics{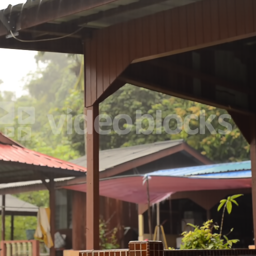}}& 
 {\includegraphics{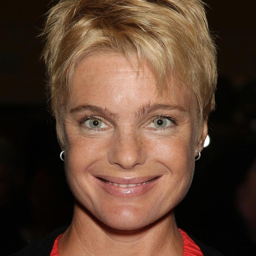}}&  {%\includegraphics{images/visuals/noo.png}
 %not available
 \raisebox{1.3cm}{
 \parbox{2cm}{\centering {\large not} \\ {\large available}}
 %\adjustbox{scale=0.25,valign=c}{ \shortstack{\text{not} \\ \text{available}}}
 }
 } \\
   \end{tabular}

\captionof{figure}{\label{fig:ood_teaser} We demonstrate \methodName on several fully blind image restoration problems (\ie, when we do not know the parametric form of the degradation model): JPEG restoration (de-artifacting), deraining, raindrop removal, and also to partially-blind restoration problems (when only the parametric form of the degradation is known but its values are unknown): deblurring,  superresolution (SR) . First row: input image. Second row: our prediction. Third row: ground-truth. \methodName is applicable to a single degraded image and does not require re-training or fine-tuning of the prior model (in our case, a diffusion model).} %Although some of the generated degraded images use Gaussian blur, \methodName recovers a generic blur kernel (without making any Gaussianity assumption). \vspace{.5em}}
\end{figure*}

To address these limitations, we propose \methodName (\methodnameacr), a novel training-free image restoration method that can handle both the fully blind (we do not know the degradation type nor its parametric form nor the values of the parameters) and partially blind degradation cases (we know the parametric form, but not the values of the parameters). %Our approach operates under complete blindness to the degradation prior, making it a practical solution especially for scenarios where the degradation is complex and difficult to model. 
We use a deterministic Denoising Diffusion Implicit Model (DDIM) as a mapping from normal noise samples to real undistorted images. Thus, given an image, we can find a corresponding input noise sample by inverting the DDIM mapping. This mapping allows us to perform image restoration by working directly in input noise space rather than in image space. Our motivation is that imposing constraints on the latent noise samples is easier, since we know very well their (ideal) probability distribution, than on the estimated images, where having an explicit model of the probability distribution is challenging. Our key observation is that noise samples that are mapped to degraded images through the deterministic DDIM lie in low probability density regions.
%The key idea is that, instead of restoring the degraded image in image space, we perform the restoration in the input space of a Denoising Diffusion Implicit Model (DDIM). Our intuition is that working directly with the manifold of natural images is more ambiguous than working with noise from a standard normal distribution, where the distribution is arguably easier to handle.
%\methodName first embeds the degraded image into the initial input noise space of a pre-trained diffusion model by applying DDIM inversion. Another key insight (or we show that )is that when the degraded image is inverted back to the DDIM input space, the resulting noise vector tends to reside in a low-density region of the standard normal distribution (depicting local correlation).
Thus, \methodName aims to estimate input noise that lies in a high probability density region (of the standard normal distribution), while generating a realistic image through the DDIM that is as close as possible to the given degraded image in the Euclidean norm (see sec.~\ref{motivation} for more details). To modify the input noise so that it satisfies the above constraints, we detect local noise patches that drifts from the standard normal distribution by using standard statistical testing and replace them with their nearest neighbors from a randomly generated set of noise vectors drawn from a standard normal distribution. Finally, we recover the restored image by applying DDIM to the modified input noise.
As we show in the experiments, \methodName is very versatile and can be applied to a wide range of degradation types, from linear and noisy processes to more complex non-linear degradations (\eg, see Figure \ref{fig:ood_teaser} in the case of old photo restoration). 

Our main contributions can be summarized as follows
\begin{itemize} \item We introduce \methodName, a new zero-shot and training-free image restoration method applicable to both fully and partially blind cases.

\item To the best of our knowledge, \methodName is the \textbf{first} method that restores an image by "restoring" its inverted noise through a diffusion model.

\item We propose a novel method for restoring the image by moving its inverted noise towards a higher density region of the standard normal distribution.

\item We achieve state-of-the-art results on various blind and partially-blind image restoration tasks, including JPEG de-artifacting, deraining, raindrop removal, motion deblurring, and super-resolution across different datasets. \end{itemize}

\begin{figure*}
    \centering
    \includegraphics[width=0.75\linewidth]{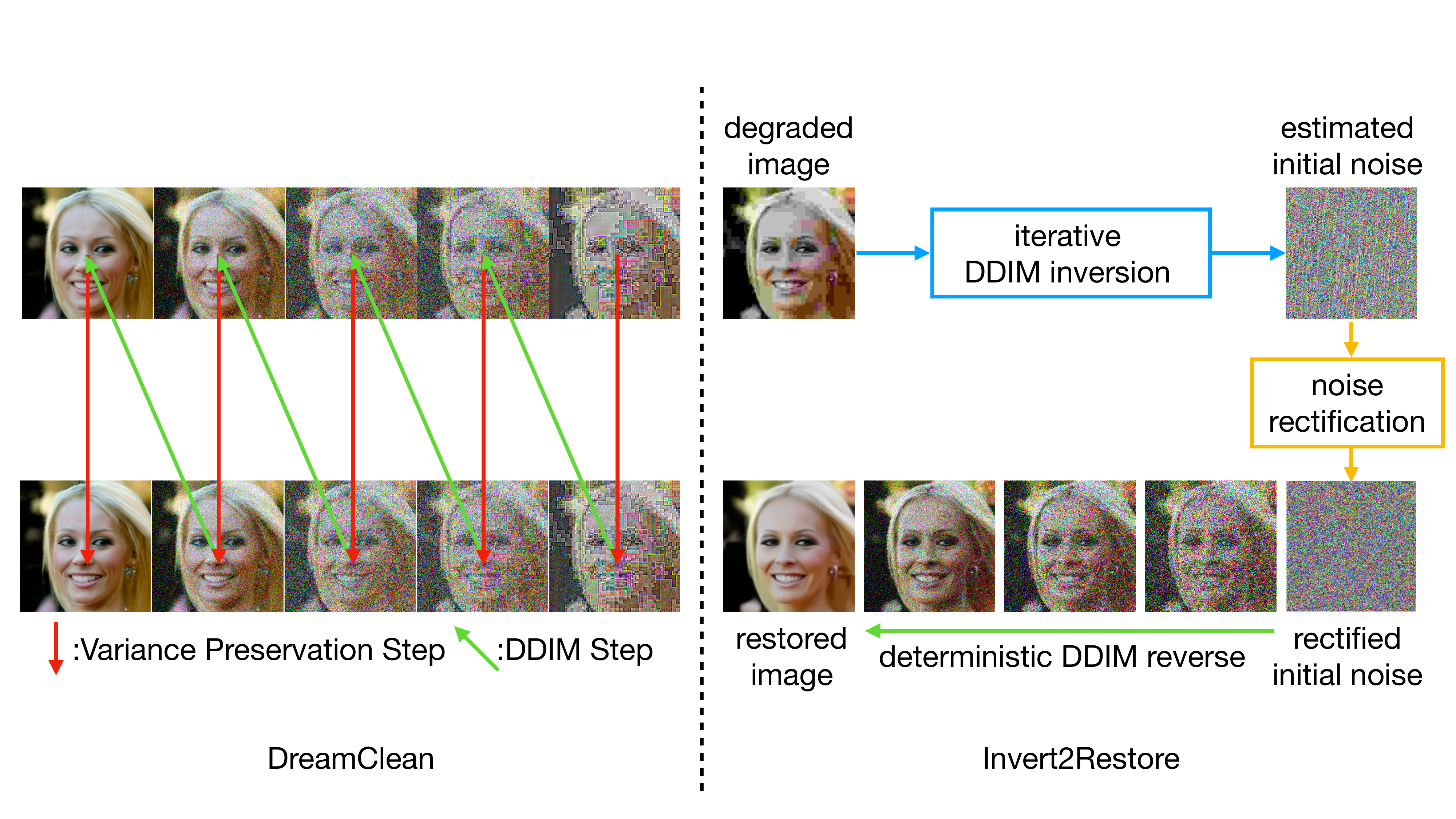}
    \caption{
    Our method vs DreamClean \cite{xiaodreamclean}. Both methods can operate in full blindness regarding the degradation model. Left: DreamClean alters the diffusion reverse sampling. In contrast, in our method we only estimate and rectify the initial noise and do not alter the reverse sampling of the diffusion. Right: A simple high-level workflow of \methodName. Our method estimates the initial noise to reconstruct the degraded image. Then, we rectify the initial noise by finding a nearby noise sample with a higher density of the standard normal distribution. Lastly, we apply the deterministic reverse diffusion to obtain the final restored image.    
    \label{fig:enter-label}
    }
\end{figure*}

% PAOLO: setting this to red (also remove the instruction at the end of the section when done)
%\begin{redsection}
\section{Related Work}

% The image restoration literature is extensive. Here, we briefly summarize related work about training-Free and zero-Shot methods with an emphasis on the diffusion-based methods, as \methodnameacr{} falls into this category. These methods operate directly on the degraded image and can be classified based on their assumptions about the degradation model.
% \subsection{Supervised Image Restoration}
% These method use a dataset with degraded-clean pairs to train a restoration model. Recent advancements in supervised image restoration have focused on leveraging diffusion models due to their ability to generate high-quality outputs through iterative refinement. One notable approach is SR3 \cite{saharia2022image}, which progressively enhances image resolution through a noise-based diffusion process trained on paired high- and low-resolution images. Palette \cite{saharia2022palette} extends diffusion models for various image-to-image translation tasks, achieving impressive results in tasks such as colorization and inpainting. Denoising Diffusion Probabilistic Models (DDPM) have also been adapted for tasks like deblurring, with models trained to reverse a learned noising process, thus gradually reconstructing clean images. DiffIR \cite{xia2023diffir} employs a diffusion-based framework to restore heavily degraded images by modeling complex noise distributions.
% \subsection{Training-Free \& Zero-Shot Image Restoration}
\noindent{\textbf{Non-Blind IR Methods.}}  
These methods assume that the degradation model is fully known  Recently, many approaches \cite{song2019generative, song2021solving, lugmayr2022repaint, kawar2022denoising, wang2022zero, chung2022diffusion} leveraging pre-trained models have been proposed. 
%For example, \cite{song2019generative} presents a method that guides the reverse diffusion process with unmasked regions to solve inpainting in a zero-shot manner. \cite{song2021solving} uses gradient guidance to address inverse problems in medical imaging. RePaint \cite{lugmayr2022repaint} tackles inpainting by guiding the diffusion process with unmasked regions. To condition the generation process, it alters the reverse diffusion iterations by sampling unmasked regions using image data. DDRM~\cite{kawar2022denoising} introduces an inverse problem solver based on posterior sampling
%, using a variational inference objective to learn the posterior distribution of the inverse problem.
%This method employs a Singular Value Decomposition to decompose degradation operators, which can pose computational challenges for high-dimensional matrices.
%DDNM~\cite{wang2022zero} proposes a zero-shot framework for IR tasks based on range-null space decomposition, by refining only the null-space content during the reverse diffusion process to ensure data consistency and realism. 
%MCG~\cite{chung2022improving} highlights the risk of stepping outside the data manifold due to iterative procedures involving reverse diffusion steps and a projection-based consistency step. They mitigate this risk using an additional correction term. 
%DPS~\cite{chung2022diffusion} offers a more general framework for handling both non-linear and noisy cases.
However, assuming that the degradation model is known limits the use of those approaches in some real-world scenarios.\\
\noindent{\textbf{Partially Blind methods. }} 
Some recent zero-shot image restoration (IR) methods \cite{fei2023generative, chung2023parallel, chihaoui2024blind, murata2023gibbsddrm} relax the assumption of fully knowing the degradation model by assuming knowledge only of its parametric form, but not its specific parameters. These methods jointly infer both the parameters of the degradation model and the clean images. BlindDPS \cite{chung2023parallel} extends the work of \cite{chung2022diffusion} to handle blind deblurring by simultaneously optimizing for both the sharp image and the blur operator through the reverse diffusion process. GibbsDDRM \cite{murata2023gibbsddrm} expands DDRM for cases where only partial knowledge of the degradation is available. BIRD \cite{chihaoui2024blind} presents a method that jointly inverts the diffusion model and infers the degradation model. In this work, we explore the inversion of diffusion models but in a different way. Unlike \cite{chihaoui2024blind}, we assume complete blindness regarding the degradation model.\\
\noindent{\textbf{Fully Blind Methods. }} 
Few works have been proposed for scenarios where the degradation model is entirely unknown or too complex to be represented in a parametric form. One pioneering approach is Deep Image Prior (DIP) ~\cite{ulyanov2018deep}, which can optimize image reconstruction without explicitly modeling the degradation. To the best of our knowledge, DreamClean~\cite{xiaodreamclean} is the only training-free and zero-shot method based on a diffusion model that works across various types of degradation while assuming no knowledge of the degradation model. DreamClean~\cite{xiaodreamclean}  operates by inverting degraded images up to an intermediate diffusion step, then by applying reverse diffusion sampling with a variance preservation step introduced after each DDIM step, iteratively refining the intermediate diffusion latent. In this work, we also address the case of fully unknown degradation models yet in a novel way. Instead of correcting the intermediate diffusion latent, our method fully inverts the degraded image back to the initial noise and carefully shifts the noise to a higher region of the standard normal distribution. Our intuition is that the input diffusion space is well-defined (standard normal distribution) and independent of the data. Once the initial noise is ``restored'', we obtain the clean image by simply applying the DDIM reverse process.

% Reset to the default section font
%\end{redsection}

\section{Image Restoration via \methodName}
We denote a clean image as \( x \sim p_x(x)\), where $p_x(x)$ is the probability density function of clean images and \( x\in \mathbb{R}^{n \times m \times c}  \) has dimensions \( n \times m \) pixels and \( c \) channels. The degradation process is modeled by a forward operator \( \mathcal{H} \) that maps the clean image \( x \) to the degraded one \( y \).  We assume that we are given a generative model to sample clean images from normal noise samples $z\sim p_z(z)$.
In practice, the generative model is a deterministic invertible and differentiable mapping $g$ induced by a pre-trained diffusion model. We assume that the pre-training is perfect, \ie, such that $p_x = g_{\#} p_z$, where  $g_{\#}$ denotes the pushforward measure induced by $g$, meaning that $p_x$ is the distribution obtained by transforming $p_z$ through $g$. By using the change of variables we have that
%$p_z(z) = p_x(g(z))\left|\nabla g(z)\right|$.
$p_x(x) = p_z(g^{-1}(x))\left|\nabla g^{-1}(x)\right|$.
Moreover, we define an image $y$ to be \emph{degraded} by a partially known or fully unknown operator $\mathcal{H}$ when $p_x(y)<\tau$, for some chosen $\tau>0$.

In this paper, we focus on the \emph{blind degradation} case, which can be categorized into two scenarios:\\  
1) \textbf{Fully blind}: The degradation model is entirely unknown, that is, we do not know what degradation it is nor its parametric form (\eg, image deraining, JPEG de-artifacting).\\  
2) \textbf{Partially-blind}: The parametric form of the degradation model is known, but its specific parameter values are not (\eg, deblurring).   

To address both scenarios, we leverage as pre-trained diffusion model the DDIM~\cite{song2020denoising}, a class of diffusion models that enables a deterministic mapping from noise space to clean images space. We thus define the deterministic mapping $g$ that transforms a standard normal noise sample $z$ into a clean image sample $x$ from the target distribution via \( x = g(z) \doteq G_{\text{DDIM}}(z) \).  
\begin{figure}[t]
\centering\
   \setkeys{Gin}{width=0.2\linewidth}
    \captionsetup[subfigure]{skip=0.0ex,
                             belowskip=0.0ex,
                             labelformat=simple}
                             \setlength{\tabcolsep}{0.1pt}
    \renewcommand\thesubfigure{}
    \small

\small
  \begin{tabular}{c@{\hspace{0.2em}}c@{\hspace{0.2em}}c}
  %\scriptsize
 % \textit{Input} & \scriptsize\textit{DIP}~\cite{chihaoui2024blind} & \scriptsize \textit{DreamClean}~\cite{xiaodreamclean} & \scriptsize \textit{Ours} & \scriptsize \textit{Target} \\
 {\includegraphics{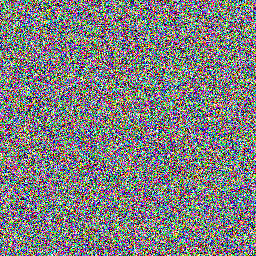}}& {\includegraphics{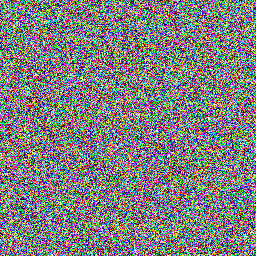}}& {\includegraphics{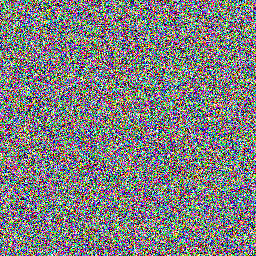}}\\
 \textit{$z_1$} & \textit{$z_2$} & \textit{$z_3$}   \\
 {\includegraphics{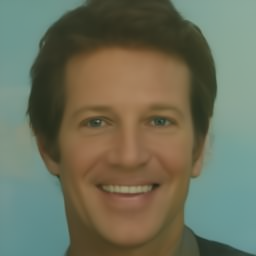}}& {\includegraphics{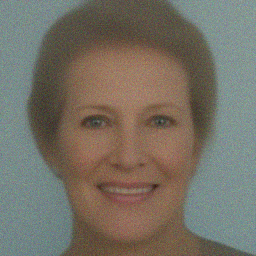}}& {\includegraphics{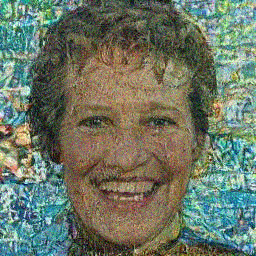}}\\
 \textit{$G_{\text{DDIM}}(z_1)$} & \textit{$G_{\text{DDIM}}(z_2)$} & \textit{$G_{\text{DDIM}}(z_3)$}   \\
   \end{tabular}

\captionof{figure}{\label{fig:motivation} Images generated by a pre-trained diffusion model (second row) fed with the respective input noise (first row). Left column: the initial noise $z_1$ is drawn from the standard normal. Middle column: $z_2 = \beta_2 z_1$, $\beta_2=0.96<1$. Right column: $z_3 = \beta_3 z_1$, $\beta_3=1.04>1$. Notice that $p_z(z_1)>p_z(z_2)$ and $p_z(z_1)>p_z(z_3)$. Thus, the quality and content of the generated images depend on whether the sampled noise is from a higher ($z_1$) or a low ($z_2$ and $z_3$) density region of \( p_z(z) =  \mathcal{N}(\mathbf{0}, \mathbf{I}) \).\vspace{.5em}}
\end{figure}

\subsection{Fully Blind Image Restoration}

We first consider the extreme case where the degradation model is completely unknown, including both its type and parametric form. %The only assumption we make is that the degradation does not entirely erase the original signal, ensuring that the restoration is possible.
%In such cases, successful restoration is only possible if the degradation retains some information from the original signal. Without this, reconstruction would be fundamentally impossible. We formalize this essential constraint in the next section.

\subsubsection{Motivation and Key Idea}\label{motivation}

In Figure~\ref{fig:motivation}, we show images generated by a mapping \( G_{\text{DDIM}} \) induced by a pre-trained diffusion model trained to map samples from \( \mathcal{N}(\mathbf{0}, \mathbf{I}) \) to human faces. %using different initial noise vectors/maps. 
As can be observed, when the input noise deviates from the standard normal distribution, the generated images become distorted and no longer resemble realistic undistorted faces (see middle and right columns in Figure~\ref{fig:motivation}, where the same noise $z_1\sim \mathcal{N}(\mathbf{0}, \mathbf{I})$ has only been scaled down to $z_2$ and up to $z_3$ respectively).  

This observation motivates us to partition the space \( \mathbb{R}^{n \times m \times c} \), in which the noise $z$ resides, into regions corresponding to likely and unlikely samples from \( \mathcal{N}(\mathbf{0}, \mathbf{I}) \). Since the standard normal distribution assigns nonzero probability density to all points in this space, we define high-density and low-density regions using a confidence level \( 1 - \alpha \), where \( \alpha \) is called the significance level and represents the probability assigned to low-density regions. A typical choice is \( \alpha = 0.05 \), corresponding to a 95\% confidence level.
In a high-dimensional space such as \( \mathbb{R}^{n \times m \times c} \), where often $n\times m \times c > 10^5$, the probability density of a standard normal distribution is largely concentrated near the surface of a hypersphere with radius \( \sqrt{mnc} \) \cite{vershynin2018random}.%, and not around zero as for scalar normal noise.  

Let \( p_z(z) \) be the probability density function of the standard normal distribution over \( \mathbb{R}^{n \times m \times c} \). The high-density region \( \mathcal{R}_H \) is defined as the set of points 
\[
\mathcal{R}_H \doteq \{ z \in \mathbb{R}^{n \times m \times c} : p_z(z) \geq \tau_\alpha \},
\]
where \( \tau_\alpha \) is a threshold chosen such that the total probability  within this region is \( 1 - \alpha \), \ie,
\[
P(z \in \mathcal{R}_H) \doteq \int_{\mathcal{R}_H} p_z(z) dz = 1 - \alpha.
\]
Similarly, the low-density region \( \mathcal{R}_L \) is defined as:
\[
\mathcal{R}_L \doteq \{ z \in \mathbb{R}^{n \times m \times c} : p_z(z) < \tau_\alpha \}.
\]
In practice, statistical normality tests can be used to determine \( \tau_\alpha \) (or equivalent statistics) automatically. A sample that fails the test is classified as belonging to \( \mathcal{R}_L \); otherwise, it belongs to \( \mathcal{R}_H \).  

% This region contains points with the lowest probability density, summing to a total probability of \( \alpha \).  while those below belong to \( \mathcal{R}_L \).  
% In this context, \( z_0 \) belongs to \( \mathcal{R}_H \) (high-density), whereas \( z_1 \) and \( z_2 \) are from \( \mathcal{R}_L \) (low-density).

\begin{proposition}
Let $y$ be a degraded image and let us define 
\begin{align}
    \tilde z = G_\text{DDIM}^{-1}(y).\label{eq:constrained_z_3}
\end{align}
%solve the following optimization problem
% \begin{align}\label{eq:constrained_z_3}
%     \tilde z = \arg \min_{z \in \mathbb{R}^{n \times m \times c}} \| y - G_\text{DDIM}(z) \|^2.
% \end{align}
Since $G_\text{DDIM}$ is invertible and differentiable, the determinant of its Jacobian admits an upper bound $B$. Then, by setting $\tau_\alpha= \tau B$, the noise $ \tilde z$ in \cref{eq:constrained_z_3} lies in a low-density region $\mathcal{R}_L$ of $\mathcal{N}(0, \mathbf{I})$.
\end{proposition}
We report the proof in Section \ref{props0} of the appendix.
The direct relationship between $\tau$ and $\tau_\alpha$ implies that the more an image is degraded, the more the corresponding $\tilde z$ lies in the tail of the normal distribution. 

As previously mentioned, we confirm experimentally in Figure~\ref{fig:motivation} that when the input noise deviates from the standard normal distribution (\ie, it belongs to $\mathcal{R}_L$), the generated image degrades. The reverse is also true: when inverting a degraded image \( y \) using a pre-trained diffusion model (trained on clean data), the resulting noise $ \tilde z = G_\text{DDIM}^{-1}(y)$ is from a low-density region of the standard normal distribution (\ie, \( \tilde{z} \in \mathcal{R}_L \)). 

%We include a formal proof in Section 1 of our supplementary material. 
Inspired by this observation, instead of restoring the degraded image directly, we take a two-step approach: 1) we first invert the degraded image by solving \cref{eq:constrained_z_3}, obtaining a noise map $\tilde z$; 2) we then refine this noise map by carefully shifting it toward a higher-density region of the standard normal distribution, \ie, toward $\mathcal{R}_H$.

\subsubsection{Solving the Optimization in \cref{eq:constrained_z_3}}

We get $ \tilde z = G_\text{DDIM}^{-1}(y)$ by solving the optimization problem
\begin{align}\label{eq:optimization}
    \tilde z = \arg \min_{z \in \mathbb{R}^{n \times m \times c}} \| y - G_\text{DDIM}(z) \|^2
\end{align}
using gradient descent. We initialize \( z \) with a random vector \( z_0 \) sampled from \( \mathcal{N}(\mathbf{0}, \mathbf{I}) \) and update it iteratively to minimize the objective function \( \| y - G_\text{DDIM}(z) \|^2 \).  
Since computing \( G_\text{DDIM}(z) \) is computationally expensive in diffusion models, we adopt the acceleration technique from \cite{chihaoui2024blind} introducing a jump term \( \delta t \) during the reverse diffusion process and thus enabling faster convergence. A detailed algorithm of the efficient computation of \( G_\text{DDIM}(z) \) is provided in Section \ref{efficient-compute} of the appendix.  
We repeat this iterative procedure for \( T \) steps until the loss reaches a sufficiently small value, yielding \( \tilde{z} \). The inversion algorithm is summarized in Algorithm~\ref{alg:iterative_solver}.  

% \begin{algorithm}[H]
%  \algsetup{linenosize=\small}
%   \scriptsize
%   \caption{ $G_\text{DDIM}( z_T, \delta t)$\label{alg:ddim}}
%   \begin{algorithmic}[1]
%   \STATE $t = T$ 
%   \WHILE{$t > 0$}  
%     \STATE $\hat{x}_{0|t} = (z_t - \sqrt{1 - \bar{\alpha}_{t}}  \epsilon_{\theta} (z_t, t))/\sqrt{\bar{\alpha}_{t}}$ 
 
%   \STATE $z_{t- {\delta t}} = \sqrt{\bar{\alpha}_{t-{\delta t}}}  \hat{x}_{0|t} + \sqrt{1 - \bar{\alpha}_{t-{\delta t}}} . \frac{z_t - \sqrt{\bar{\alpha}_{t}} \hat{x}_{0|t}}{\sqrt{1 - \bar{\alpha}_{t}}}$
 
%    \STATE $t \leftarrow t - \delta t$
%   \ENDWHILE\\
%  \STATE return $\hat x_{0}$
% \end{algorithmic}

% \end{algorithm}

\begin{algorithm}[t]
 \algsetup{linenosize=\small}
  %\scriptsize
  \caption{Iterative Solver for Problem~\eqref{eq:optimization} \label{alg:iterative_solver}}
  \begin{algorithmic}[1]
  \REQUIRE Degraded image $y$, learning rate $\eta$, $G_\text{DDIM}$
  \ENSURE Return $\tilde {z}$\\

 \STATE Initialize $z^0  \sim \mathcal{N}(\mathbf{0}, \mathbf{I})$ 
 \FOR{$k: 0 \rightarrow N-1$} 

  %\WHILE{$k:1\rightarrow N$ and $\|G_\text{DDIM}( z^k) - y\| > \varepsilon$}  
  \STATE $z^{k+1} = z^{k} - \eta \nabla_{z} \|G_\text{DDIM}( z^k) - y\|^2$
  
  \ENDFOR\\
  \STATE \textbf{return} $\tilde {z}= z^{N}$
\end{algorithmic}

\end{algorithm}

\newcommand{\simpletilde}[1]{\hspace{0pt}\mathaccent"7E{#1}}
\subsubsection{Moving $\simpletilde{z}$ to a Higher Density Region}% of the standard normal distribution}

We cast the problem of finding a generated image that is realistic, artifact-free and as close as possible to the given degraded image $y$, as that of finding the noise map \( z^{*} = \arg \min_{z \in \mathcal{R}_H} \| z - \tilde{z} \| \). Since
we are not aware of an exact solution to this problem, we propose a sampling-based solver. %since \( \tilde{z} \) is typically high-dimensional, a direct search would be computationally inefficient.  
We adopt a local approach based on the intuition that a noise map is a sample from the high density region of the standard normal if all its local patches also are samples of the high density regions of the corresponding (lower-dimensional) standard normal distribution. Thus, we first identify local patches that deviate from normality using a statistical test. Then, we correct these patches by carefully replacing them with patches sampled from a standard normal distribution.  

\noindent\textbf{Detecting Defective Local Patches of $\simpletilde{z}$.} 
Our goal is to identify local patches of \( \tilde{z} \) that are unlikely to be samples from \( \mathcal{N}(\mathbf{0}, \mathbf{I}) \). To achieve this, we process \( \tilde{z} \) using a sliding window approach, where each patch has dimensions \( k \times k \times c \), with \( k \) representing the spatial window size (\eg, \( k = 4 \)) and \( c \) the number of channels. Each patch is then flattened into a 1D vector of size \( m = ck^2 \).  
We apply classical statistical hypothesis testing to determine whether each flattened patch follows the standard normal distribution. Given a confidence level of \( 1 - \alpha \), the test either rejects or fails to reject the null hypothesis that the patch is drawn from \( \mathcal{N}(\mathbf{0}, \mathbf{I}) \).  
After testing all patches, we construct a binary mask \( \mathbf{M} \in \{0,1\}^{n \times m \times c} \), where \( \mathbf{M}(x, y, v) = 1 \) if the noise at location \( (x, y) \) of the $v$-th channel belongs to a patch that fails the normality test.

\noindent\textbf{Enhancing $\tilde{z}$ through Sampling and Substitution.} 
To shift \( \tilde{z} \) toward a higher-density region of \( \mathcal{N}(\mathbf{0}, \mathbf{I}) \), we replace local patches where at least one pixel \((x, y)\) satisfies \(\mathbf{M}(x, y, v) = 1\) with patches sampled from \( \mathcal{N}(\mathbf{0}, \mathbf{I}) \). However, substituting these patches with purely random samples may generate a clean image that does not preserve the original content of \( y \), potentially leading to a perceptually valid but unrelated image instead of a faithful reconstruction of the clean image \( x \). 
To address this, we divide \( \tilde{z} \) into non-overlapping adjacent tiles \( \tilde{z}_{i,j,v} \) of size \( k \times k \). For each tile \( \tilde{z}_{i,j,v} \) that fails the normality test—where \( i,j \) index the spatial coordinates of the tile center and \( v \) denotes the corresponding channel—we generate a set \( Z \) of \( S \) random samples of size \( k \times k \) from \( \mathcal{N}(\mathbf{0}, \mathbf{I}) \) (\eg, \( S = 5 \times 10^4 \)).  

We then replace each defective tile \( \tilde{z}_{i,j,v} \) with the closest match \( z^{\text{sample}}_{i,j,v} \) from \( Z \), where the optimal replacement is determined as follows
\begin{align}
z^{\text{sample}}_{i,j,v} = \arg \min_{z^{\text{sample}} \in Z } \| \tilde{z}_{i,j,v} - z^{\text{sample}} \|^2.
\label{eq:samplez}    
\end{align}

We define $z^{\text{sample}} \in \mathbb{R}^{n \times m \times c}$ by collecting all the tiles $z^{\text{sample}}_{i,j,v}$ and by rearranging them as in the original $\tilde{z}$ tensor. Finally, we define  $z^{*}$ as the combination of $\tilde z$ and $z^{\text{sample}}$ based on the mask $M$
\begin{align}\label{eq:constrained_z_6}
    z^{*} = (\mathds{1} - \mathbf{M} ) \odot \tilde z + \mathbf{M} \odot  z^{\text{sample}}
\end{align}
where $\mathds{1}$ is the tensor of all ones of the same size as $\tilde z$. 

\subsection{Partially Blind Image Restoration}

The key difference from the fully blind case is that the parametric form of the degradation \( \mathcal{H} \) is known,  parameterized by some parameter vector \( \theta \), and denoted as \( \mathcal{H}_\theta \). This allows us to leverage \( \mathcal{H}_\theta \) to obtain the inverted noise \( \tilde{z} \) more directly. Specifically, in this case, we aim to solve the following optimization problem to estimate \( \tilde{z} \):
\begin{align}\label{eq:constrained_z_10}
\tilde{z}, \theta^* = \arg \min_{z \in \mathbb{R}^{n \times m \times c}, \theta} \| y - \mathcal{H}_\theta(G_\text{DDIM}(z)) \|^2.
\end{align}
We solve this problem using gradient descent, similar to Problem~\eqref{eq:optimization}. However, this time, we jointly optimize both \( z \) and \( \theta \). The process starts with random initialization of \( z \) and \( \theta \), followed by iterative gradient updates on both variables.  
%Due to limited space, we provide the detailed algorithm in our supplementary material.

At the end of the optimization, we are solely interested in the final estimate of \( \tilde{z} \). However, also in this case there is no guarantee that \( \tilde{z} \) lies in a high-density region of the standard normal distribution. Therefore, we proceed as in the fully blind case and apply the approach outlined in Section 3.1.3 to refine \( \tilde{z} \).

%\subsection{Realism via Ensuring Local Gaussianity}
\begin{table*}[t]
\centering
\caption{ Quantitative comparison with \textbf{partially blind zero-shot} methods on several image restoration tasks where the \textbf{parametric form} of the degradation is  \textbf{known} on the CelebA validation dataset. The best method is indicated in bold.\label{tab:partially-blind}}
     \centering
    \footnotesize
    \captionsetup{font=tiny}
%\vspace{-0.3cm}
\begin{tabularx}{\textwidth}
{@{\extracolsep{\fill}}lccccccccc
%@{\hspace{.5em}}c@{\hspace{0.5em}}c@{\hspace{0.5em}}c@{\hspace{1.em}}c@{\hspace{0.5em}}c@{\hspace{0.5em}}c@{\hspace{1.em}}c@{\hspace{0.5em}}c@{\hspace{0.5em}}c@{\hspace{0.5em}}
}
\toprule
Method & \multicolumn{3}{c}{Motion Deblur}  & \multicolumn{3}{c}{Gaussian Deblur}  & \multicolumn{3}{c}{ $8 \times$ SR}\\ 
%\cline{3-6}
%\vspace{0.1cm}
     & PSNR $\uparrow$ & SSIM $\uparrow$  & LPIPS $\downarrow$  & PSNR $\uparrow$ & SSIM $\uparrow$  & LPIPS $\downarrow$ & PSNR $\uparrow$  & SSIM $\uparrow$ & LPIPS $\downarrow$\\ \midrule
     
    DIP~\cite{ulyanov2018deep}      &    20.34     & 0.451         &  0.524 & 20.82    &  0.508 & 0.472      &  20.46 & 0.494  & 0.487           \\
     GDP~\cite{fei2023generative}   & {21.74}   & {0.612}       &  {0.387} & {22.31}       &  {0.635} & {0.374}      &  21.79 & 0.620 & 0.379      \\  
   
    BIRD~\cite{chihaoui2024blind}    & {22.32}   & {0.651}       &  {0.359} & {22.54}       &  {0.641} & {0.365}      &  \textbf{22.73}& {0.634} & {0.347}      \\     
     
        Gibbsddrm~\cite{murata2023gibbsddrm}      & {21.42}   & {0.592}       &  {0.394} & {22.04}       &  {0.623} & {0.377}      &  {21.65} & {0.603} & {0.355}      \\ 
          BlindDPS~\cite{chung2023parallel}   & {22.07}   & {0.624}       &  {0.368}  & {22.64}       &  {0.645} & {0.354}      &  21.94 & 0.611 & 0.368      \\ 
      Ours (\methodnameacr)    & \textbf{22.56}   & \textbf{0.666}       &  \textbf{0.347} & \textbf{23.02}       &  \textbf{0.664} & \textbf{0.348}      &  {22.62} & \textbf{0.671} & \textbf{0.330}     \\  \bottomrule 
\end{tabularx}
\end{table*}

\subsection{\methodName}

Given a degraded image \( y \), we first invert it to obtain \( \tilde{z} \). In the fully blind case, we solve the optimization Problem~\eqref{eq:optimization}, while in the partially-blind case, we solve the optimization Problem~\eqref{eq:constrained_z_10}. Next, we detect all local patches of \( \tilde{z} \) that deviate from the standard normal distribution by constructing a binary mask \( \mathbf{M} \) using a standard statistical test for normality. For the patches that originate from low-density regions of the standard normal distribution, we replace them with nearest-neighbor samples drawn from \( \mathcal{N}(\mathbf{0}, \mathbf{I}) \). Finally, we compute \( z^{*} \) using  \cref{eq:constrained_z_6}. The final restored image is obtained as \( \hat{x} = G_\text{DDIM}(z^{*}) \). This entire procedure, which we call \methodName (\methodnameacr), is summarized in \cref{alg:i2r}.

\begin{algorithm}[t!]
 \algsetup{linenosize=\small}
  %\scriptsize
  \caption{\methodnameacr{}\label{alg:i2r}}
  \begin{algorithmic}[1]
  \REQUIRE Degraded image $y$, $G_\text{DDIM}$ 
  \ENSURE Restored image $\hat{x}$\\
  \IF {If the degradation is fully unknown}
    \STATE {Compute $\tilde{z}$ by solving Problem~\eqref{eq:optimization}}
\ELSE
    \STATE {Compute $\tilde{z}$ by solving Problem~\eqref{eq:constrained_z_10}}
\ENDIF
  \STATE Compute the binary mask $\mathbf{M}$ 
  \STATE Select $z^{\text{sample}}$ by solving Problem~\eqref{eq:samplez} through search
    \STATE Compute $z^{*}$ by using \cref{eq:constrained_z_6} 
  \STATE Return the restored image $\hat{x}=G_\text{DDIM}(z^{*})$\\
\end{algorithmic}

\end{algorithm}

%%% todo
    
\section{Experiments}

\begin{figure*}[t]
\centering\
   \setkeys{Gin}{width=0.12\linewidth}
    \captionsetup[subfigure]{skip=0.0ex,
                             belowskip=0.0ex,
                             labelformat=simple}
                             \setlength{\tabcolsep}{1.5pt}
    \renewcommand\thesubfigure{}
    \small

\small
  \begin{tabular}{cccccc}
  \textit{Input} & \textit{BlindDPS}~\cite{chung2023parallel}  & \textit{BIRD}~\cite{chihaoui2024blind} & \textit{DreamClean}~\cite{xiaodreamclean} & \textit{Ours (\methodnameacr)} & \textit{Target} \\
 {\includegraphics{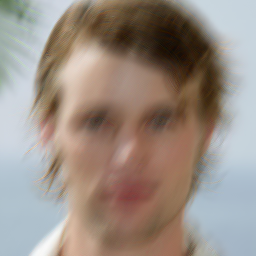}}& {\includegraphics{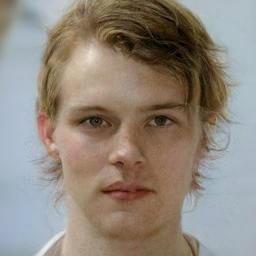}}& {\includegraphics{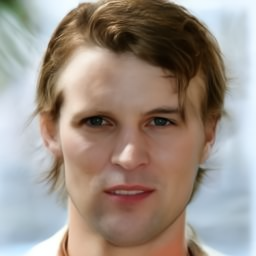}}& {\includegraphics{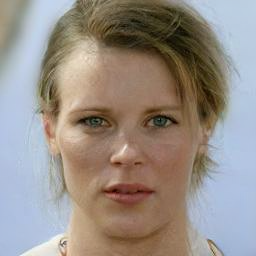}} & 
 {\includegraphics{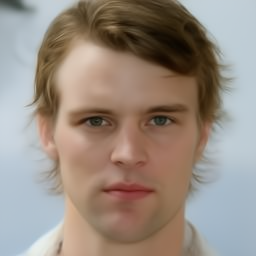}} & 
 {\includegraphics{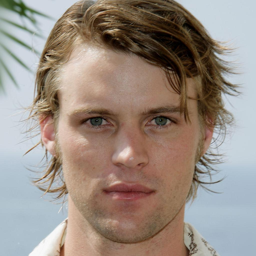}}\\
  {\includegraphics{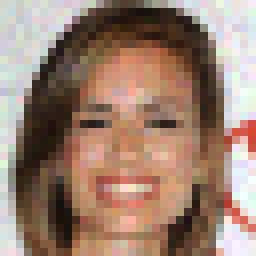}}& {\includegraphics{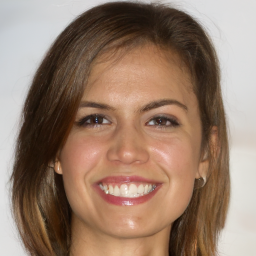}}& {\includegraphics{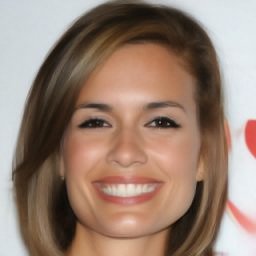}}& {\includegraphics{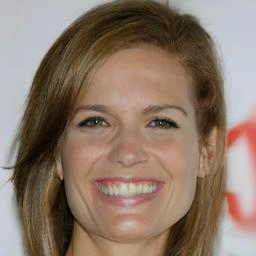}} & 
 {\includegraphics{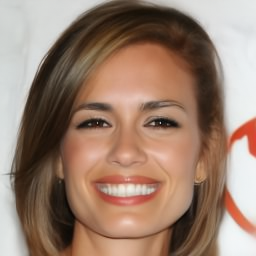}} & 
 {\includegraphics{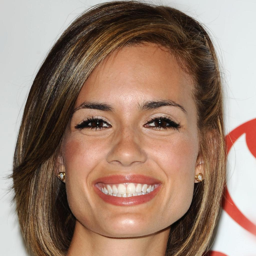}}\\
 % {\includegraphics{images/visuals/input_jpeg1.png}}& {\includegraphics{images/visuals/noo.png}}& {\includegraphics{images/visuals/noo.png}}& {\includegraphics{images/visuals/dreamclean_jpeg1.png}} & 
% {\includegraphics{images/visuals/ours_jpeg1.png}} & 
% {\includegraphics{images/visuals/target_jpeg1.png}}\\
   \end{tabular}

\captionof{figure}{\label{fig:celeba} Qualitative comparisons of different \textbf{partially blind} restoration tasks on CelebA (the degradation operator is known, but not its parameters). First row: motion deblurring. Second row: $4\times$ super-resolution.\vspace{.5em}}
\end{figure*}
\noindent\textbf{Experimental Settings. }
We validate the effectiveness of Invert2Restore using unconditional diffusion models trained on CelebA~\cite{liu2018large}, ImageNet~\cite{deng2009imagenet} where the backbone is a Unet~\cite{ronneberger2015u} architecture. To ensure a fair comparison, we use the same pre-trained models across all competing methods.  
For quantitative evaluation, we conduct experiments on classic image restoration tasks, including structured degradations (\eg, Gaussian deblurring, super-resolution) and complex unstructured degradations (\eg, deraining, JPEG artifact removal). Performance is measured using Peak Signal-to-Noise Ratio (PSNR) and Structural Similarity Index Measure (SSIM) for fidelity assessment, while Learned Perceptual Image Patch Similarity (LPIPS) is used as a perceptual metric.  
We evaluate our method on ImageNet 1K, CelebA 1K, and the validation set of SPA~\cite{li2022toward} for real-world deraining, using images of size \(256 \times 256\) pixels. Our approach is compared against state-of-the-art zero-shot and partially-blind methods, including GDP~\cite{fei2023generative}, BIRD~\cite{chihaoui2024blind}, GibbsDDRM~\cite{murata2023gibbsddrm}, and BlindDPS~\cite{chung2023parallel}. Additionally, we compare against fully blind zero-shot methods such as DIP~\cite{ulyanov2018deep} and the recent state-of-the-art DreamClean~\cite{xiaodreamclean}. Notably, only a few methods have been proposed for fully blind restoration tasks (\eg, deraining), which explains the smaller number of methods compared in Table~\ref{tab:fully-blind} relative to Table~\ref{tab:partially-blind}.  
For Gaussian blur and motion blur, we generate blurry inputs using a \(41 \times 41\) Gaussian kernel and motion kernel, respectively. For super-resolution, we apply an \(8 \times 8\) Gaussian kernel followed by an \(8\times\) downsampling operation. For JPEG artifact removal, we generate degraded inputs using the Imageio~\cite{chandra2001imageio} Python library with a quality factor of \(q=5\). In all tasks, we consider the presence of additive noise with a standard deviation of \(\sigma=0.03\). For raindrop removal, we use an online tool\footnote{https://funny.pho.to/rain-drops-effect/} to generate degraded inputs.  
Unless otherwise stated, we set \(\alpha = 0.05\) (95\% confidence level), use \(N = 150\) iterations for the iterative solver in Algorithm 1, set \(k = 4\), and use \(S = 50{,}000\) samples. Optimization is performed using the Adam optimizer with a learning rate of 0.001. We use the same feed-forward network in \cite{ren2020neural} for our blur
kernel estimation in the case of image deblurring and superresolution.  For normality testing, we employ D’Agostino and Pearson’s normality test~\cite{diagostino1971omnibus}. We include a review on  D’Agostino and Pearson’s normality test in Section \ref{statistical-testing} of the appendix.

\begin{figure}[t]
\centering\
   \setkeys{Gin}{width=0.2\linewidth}
    \captionsetup[subfigure]{skip=0.0ex,
                             belowskip=0.0ex,
                             labelformat=simple}
                             \setlength{\tabcolsep}{0.0pt}
    \renewcommand\thesubfigure{}
    \small

\small
  \begin{tabular}{ccccc}
  \scriptsize
  \textit{Input} & \scriptsize\textit{DIP}~\cite{chihaoui2024blind} & \scriptsize \textit{DreamClean}~\cite{xiaodreamclean} & \scriptsize \textit{Ours  (\methodnameacr)} & \scriptsize \textit{Target} \\
 {\includegraphics{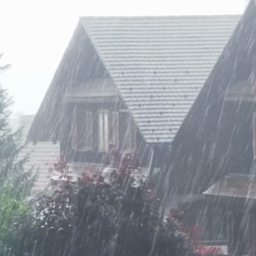}}& {\includegraphics{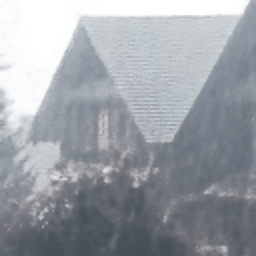}}& {\includegraphics{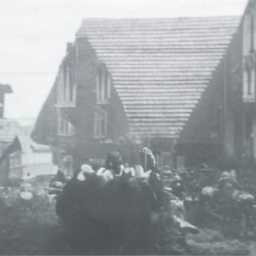}}& {\includegraphics{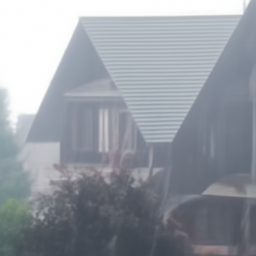}} & 
 {\includegraphics{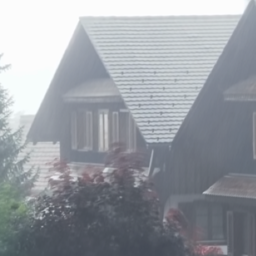}}\\
 {\includegraphics{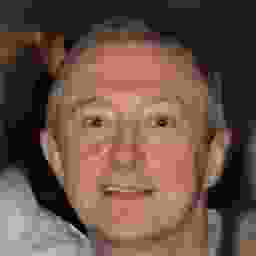}}& {\includegraphics{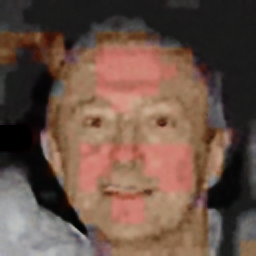}}& {\includegraphics{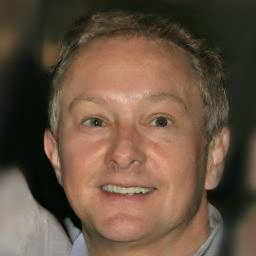}}& {\includegraphics{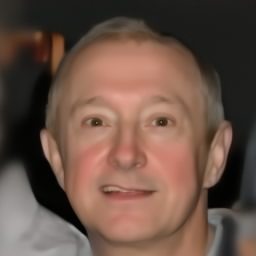}} & 
 {\includegraphics{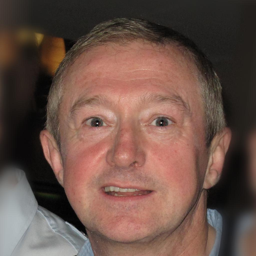}} \\ {\includegraphics{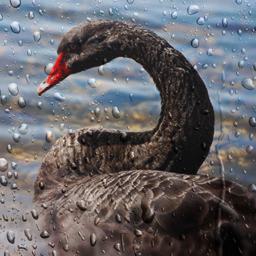}}& {\includegraphics{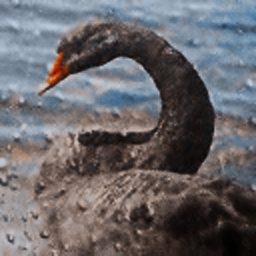}}& {\includegraphics{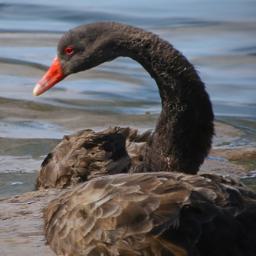}}& {\includegraphics{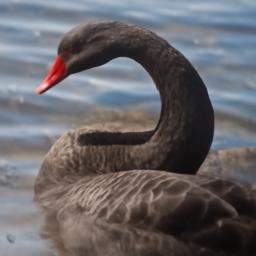}} & 
 {\includegraphics{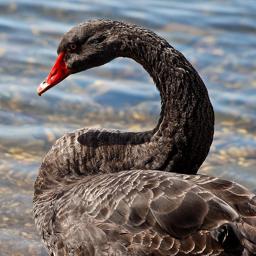}}\\ 
 % {\includegraphics{images/visuals/input_jpeg1.png}}& {\includegraphics{images/visuals/noo.png}}& {\includegraphics{images/visuals/noo.png}}& {\includegraphics{images/visuals/dreamclean_jpeg1.png}} & 
% {\includegraphics{images/visuals/ours_jpeg1.png}} & 
% {\includegraphics{images/visuals/target_jpeg1.png}}\\
   \end{tabular}

\captionof{figure}{\label{fig:full-blind} Qualitative comparisons of different restoration tasks where the parametric form of the degradation is unknown. First row: Deraining. Second row: JPEG de-artifacting. Third row: raindrop removal.\vspace{.5em}}
\end{figure}

\begin{figure}[t]
\centering\
   \setkeys{Gin}{width=0.16\linewidth}
    \captionsetup[subfigure]{skip=0.0ex,
                             belowskip=0.0ex,
                             labelformat=simple}
                             \setlength{\tabcolsep}{0.3pt}
    \renewcommand\thesubfigure{}
    \small

\small
  \begin{tabular}{cccccc}
  
 {\includegraphics{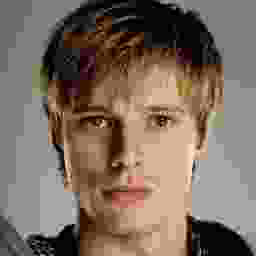}}& {\includegraphics{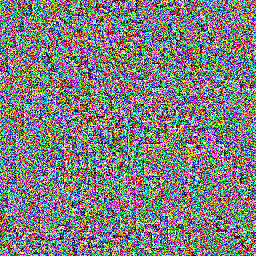}}& {\includegraphics{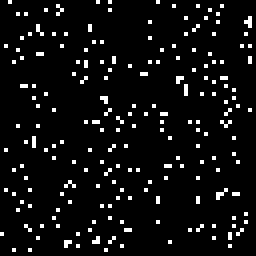}}&{\includegraphics{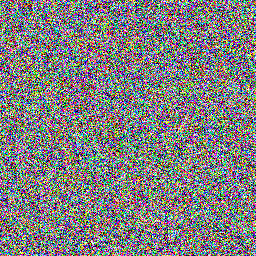}} & 
 {\includegraphics{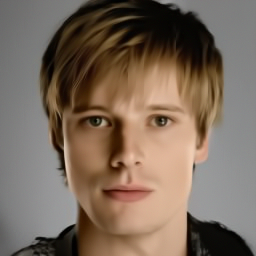}}& {\includegraphics{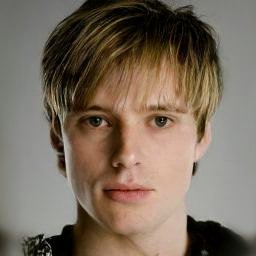}}\\
  \scriptsize
  \textit{Input} & \scriptsize\textit{$\tilde{z}$} & \scriptsize \textit{mask $\mathbf{M}$}&
   \scriptsize \textit{$z^*$} & \scriptsize \textit{Ours} & \scriptsize \textit{ground truth}  \\

   \end{tabular}

\captionof{figure}{\label{fig:refinement} From left to right: input image corrupted with JPEG artifacts; inverted noise $\tilde{z}$; binary mask $\mathbf{M}$ (the while pixels denote the locations where noise elements should be substituted); rectified noise $z^*$; restored image; ground-truth.}% \vspace{.5em}}
\end{figure}

\noindent\textbf{Quantitative and Qualitative Comparison.}
Tables~\ref{tab:partially-blind} and \ref{tab:fully-blind} present the performance of \methodName alongside other compared methods in both the partially blind and fully blind settings. \methodName consistently outperforms or is on par with state-of-the-art approaches across various image restoration tasks. Notably, methods such as \cite{fei2023generative, chihaoui2024blind, murata2023gibbsddrm, chung2023parallel} are not applicable in the fully blind case, as they require knowledge of the parametric form of the degradation model. Therefore, they are not included in Table \ref{tab:fully-blind}.
Figure~\ref{fig:celeba} presents a visual comparison of CelebA images restored for motion deblurring and super-resolution tasks.
Figure~\ref{fig:full-blind} illustrates a visual comparison in the fully blind setting. Despite having no prior knowledge of the degradation, \methodName produces plausible reconstructions and tends to preserve higher fidelity compared to other competing methods. In Figure~\ref{fig:refinement}, we visualize the intermediate steps of \methodName: the inverted noise \(\tilde{z}\), the binary mask \(M\), the rectified noise \(z^*\), and the final restored output.

\begin{table*}[t]
\centering
\caption{Quantitative comparison with \textbf{fully blind zero-shot} methods on several image restoration tasks where the \textbf{parametric form} of the degradation is  \textbf{unknown}. The best method is indicated in bold.\label{tab:fully-blind}}
     \centering
    \footnotesize
    \captionsetup{font=tiny}
%\vspace{-0.3cm}
\begin{tabular*}{\textwidth}{@{\extracolsep{\fill}}lccccccccr
%@{\hspace{.0em}}l@{\hspace{.5em}}c@{\hspace{0.5em}}c@{\hspace{0.5em}}c@{\hspace{1.em}}c@{\hspace{0.5em}}c@{\hspace{0.5em}}c@{\hspace{1.em}}c@{\hspace{0.5em}}c@{\hspace{0.5em}}c@{\hspace{0.5em}}c@{\hspace{0.5em}}c@{\hspace{0.5em}}c@{\hspace{0.5em}}
}
\toprule
Method & \multicolumn{3}{c}{Deraining}  & \multicolumn{3}{c}{Raindrop removal}& \multicolumn{3}{c}{JPEG De-artifacting}\\ 
%\cline{3-6}
%\vspace{0.1cm}
     & PSNR $\uparrow$ & SSIM $\uparrow$  & LPIPS $\downarrow$  & PSNR $\uparrow$ & SSIM $\uparrow$  & LPIPS $\downarrow$ & PSNR $\uparrow$ & SSIM $\uparrow$  & LPIPS $\downarrow$\\ \midrule
       DIP~\cite{ulyanov2018deep}      &   19.23    & 0.483         &  0.635 &  20.12    & 0.493         &  0.604 & 21.25    &  0.589 & 0.417           \\
     DreamClean~\cite{xiaodreamclean}      & {23.89}   & {0.624}       &  {0.516} & {22.43}       &  {0.547} & {0.491}  & 22.48    &  0.634 & \textbf{0.327}         \\ 
      Ours (\methodnameacr)    & \textbf{27.12}   & \textbf{0.809}       &  \textbf{0.355} &  \textbf{22.67} & \textbf{0.569}    & \textbf{0.483}           & \textbf{25.07}    &  \textbf{0.757} & 0.340   \\  \bottomrule 
\end{tabular*}
\end{table*}

\section{Ablations } 

We conduct different ablations to analyze the impact of each component of \methodnameacr.

\begin{table}[t]
  
\centering
\caption{Effect of the number of random samples $S$ on  \methodName on the CelebA validation dataset. PSNR (dB) is reported when using  different $S$ values. \label{tab:N}}
     \centering
    \footnotesize
    \captionsetup{font=scriptsize}
%\vspace{-0.3cm}
\begin{tabular*}{\linewidth}{
@{\extracolsep{\fill}}lcc
%@{\hspace{0em}}l
%@{\hspace{3em}}c@{\hspace{2em}}c
%@{\hspace{0em}}
}
\toprule
$S$     & JPEG-deartifacting & Superresolution  \\ \midrule
 % DIP~\cite{ulyanov2018deep}     &     520      &     1.2           \\
$2\cdot 10^4$    & 24.93 &  22.51\\
$5\cdot 10^4$    &      25.07 & 22.62            \\
$10^5$    &      25.10 & 22.56           \\

\bottomrule
\end{tabular*}
\end{table}

\begin{table}[t]
  
\centering
\caption{Effect of the patch size $k$ on  \methodName performance. PSNR (dB) is reported for different $k$ values. \label{tab:k}}
     \centering
    \footnotesize
    \captionsetup{font=scriptsize}
%\vspace{-0.3cm}
\begin{tabular*}{\linewidth}{
@{\extracolsep{\fill}}lcc
%@{\hspace{0em}}l@{\hspace{3em}}c@{\hspace{5em}}c@{\hspace{0em}}
}
\toprule
$k$     & JPEG-deartifacting & Gaussian deblurring  \\ \midrule
 % DIP~\cite{ulyanov2018deep}     &     520      &     1.2           \\
$4$    & 25.07 & 23.02\\
$6$    &      24.92 & 22.95            \\
$8$    &      24.71 & 22.83           \\

\bottomrule
\end{tabular*}
\end{table}

\begin{table}[t]
  
\centering
\caption{Effect of our neraest neighbors substitution scheme on  \methodName performance. PSNR (dB) is reported . \label{tab:neraest-neighbors}}
     \centering
    \footnotesize
    \captionsetup{font=scriptsize}
%\vspace{-0.3cm}
\begin{tabular*}{\linewidth}{
@{\extracolsep{\fill}}lcc
%@{\hspace{0em}}l@{\hspace{1em}}c@{\hspace{1em}}c@{\hspace{0em}}
}
\toprule
Rectification method     & JPEG-deartifacting & Motion deblurring  \\ \midrule
 % DIP~\cite{ulyanov2018deep}     &     520      &     1.2           \\
No $\tilde{z}$  rectification  & 23.26 & 21.74\\
Random substitution    & 22.13 & 20.31\\
\methodName   &      25.07 & 22.56            \\

\bottomrule
\end{tabular*}
\end{table}

\begin{figure}[t]
\centering\
   \setkeys{Gin}{width=0.2\linewidth}
    \captionsetup[subfigure]{skip=0.0ex,
                             belowskip=0.0ex,
                             labelformat=simple}
                             \setlength{\tabcolsep}{0.3pt}
    \renewcommand\thesubfigure{}
    \small

\small
  \begin{tabular}{ccccc}
  
 {\includegraphics{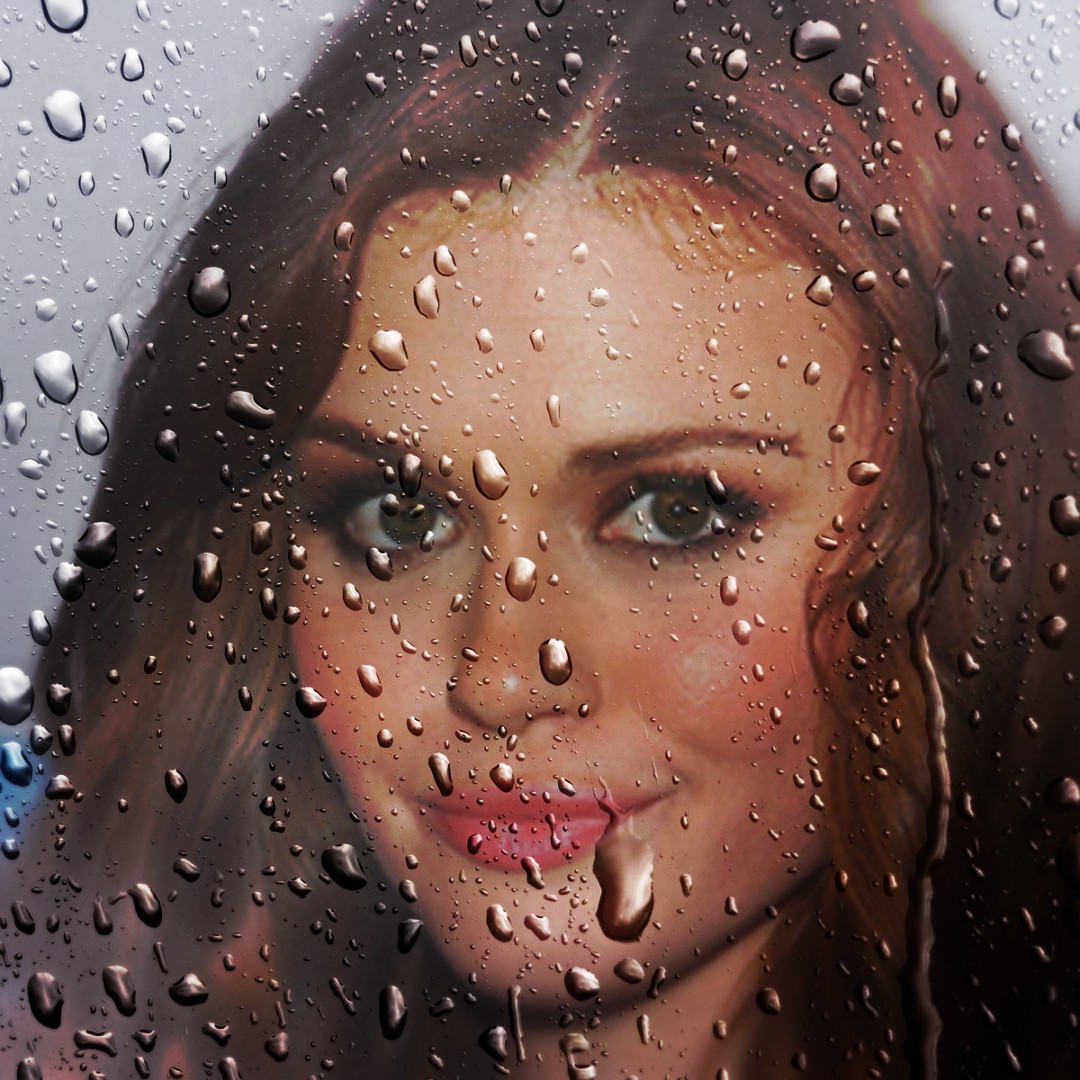}}& {\includegraphics{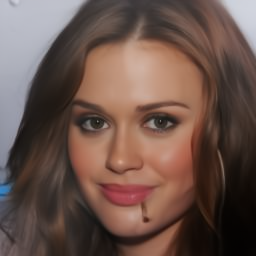}}& {\includegraphics{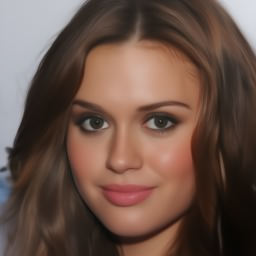}}&{\includegraphics{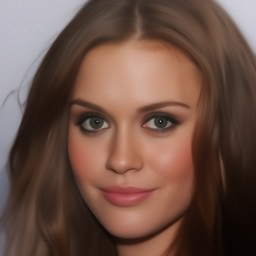}} & 
 {\includegraphics{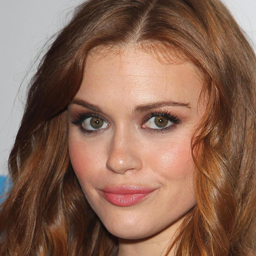}}\\
  \scriptsize
  \textit{Input} & \scriptsize\textit{Ours(\tiny$\alpha=0.01$)} & \scriptsize \textit{Ours(\tiny$\alpha=0.05$)} &  \scriptsize \textit{Ours(\tiny$\alpha=0.1$)}&\scriptsize \textit{Ground-truth}  \\

   \end{tabular}

\captionof{figure}{\label{fig:alpha} \methodnameacr outputs based on different $\alpha$ values.}% \vspace{.5em}}
\end{figure}

\noindent\textbf{Effect of the significance level $\alpha$.}
In Table \ref{tab:alpha}, we show the effect of the significance level $\alpha$ during the statistical testing of normality.  $\alpha=0.05$ yields the best performance achieving a balance between fidelity and realism. Figure~\ref{fig:alpha} shows \methodnameacr output based on different $\alpha$ values.

\begin{figure}[t]
\centering\
   \setkeys{Gin}{width=0.27\linewidth}
    \captionsetup[subfigure]{skip=0.0ex,
                             belowskip=0.0ex,
                             labelformat=simple}
                             \setlength{\tabcolsep}{0.3pt}
    \renewcommand\thesubfigure{}
    \small

\small
  \begin{tabular}{ccc}
  
 {\includegraphics{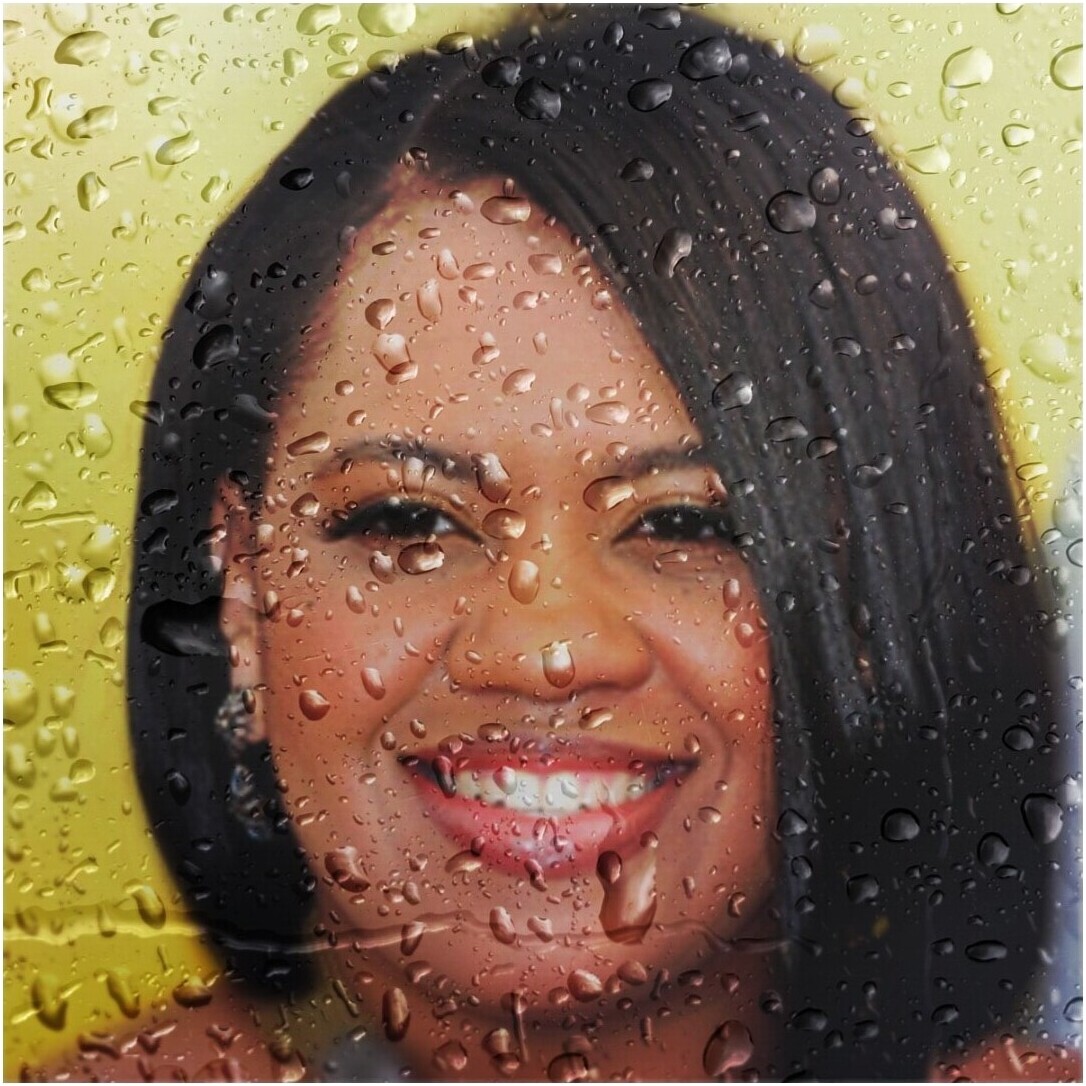}}& {\includegraphics{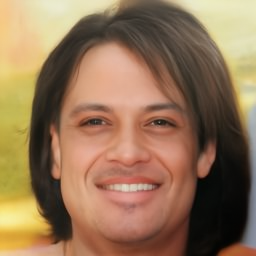}}& {\includegraphics{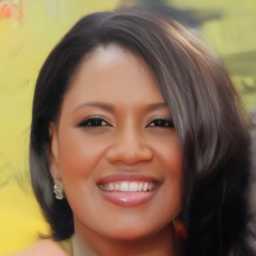}}\\
  \scriptsize
  \textit{Input} & \scriptsize  \textit{Random substitution} & \scriptsize  \textit{Random substitution}\\
  {\includegraphics{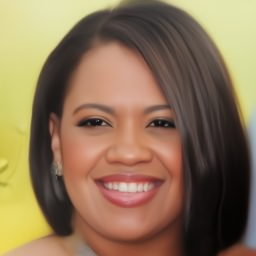}} & 
 {\includegraphics{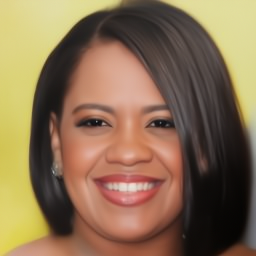}}& {\includegraphics{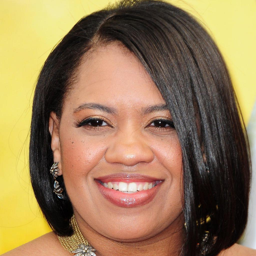}}\\
    \scriptsize \textit{Ours (\methodnameacr) sample \#1} & \scriptsize \textit{Ours (\methodnameacr) sample \#2} & \scriptsize \textit{Ground-truth}  \\

   \end{tabular}

\captionof{figure}{\label{fig:substitution-scheme} Utility of noise \(\tilde{z}\) refinement using our nearest neighbors scheme. First row: 
Degraded input followed by two predictions with different initializations, where noise substitution is performed randomly. Second row:  
Two predictions by \methodName\ with different initializations, obtained by substituting noise using the nearest neighbors scheme.  
Our scheme maintains higher fidelity to the original input.}
\end{figure}

\noindent\textbf{Effect of noise map $\tilde{z}$ refinement.}
Table \ref{tab:neraest-neighbors} compares performance under three different strategies: (1) no rectification of \( \tilde{z} \), (2) a baseline approach where defective local patches are replaced with random noise patches from $\mathcal{N}(\mathbf{0}, \mathbf{I})$, and (3) our method, where we sample \( S \) noise patches and select the nearest neighbor for substitution. Our approach achieves the best performance and better preserves fidelity to the original image, as also illustrated in Figure \ref{fig:substitution-scheme}.  

\noindent\textbf{Effect of the size of set $Z$, $S$.}
Table \ref{tab:N} illustrates the effect of the number of randomly generated samples used for substitution during noise map refinement. \methodnameacr remains robust for \(S \) in the range \([2 \times 10^4, 10^5]\). Setting \( S = 5 \times 10^4 \) provides a good balance between image quality and computational efficiency.

\noindent\textbf{Effect of the patch size $k$.}
Table \ref{tab:k} shows the effect of patch size \( k \) on performance. A smaller patch size (\( k=4 \)) tends to yield slightly better results. The reported results are based on a fixed set size \( S = 5 \times 10^4 \).

\begin{table}[]
  
\centering
\caption{Effect the significance level $\alpha$ on \methodName performance. PSNR (dB) is reported when using different $\alpha$ values.  \label{tab:alpha}}
     \centering
    \footnotesize
    \captionsetup{font=scriptsize}
%\vspace{-0.3cm}
\begin{tabular*}{\linewidth}{
@{\extracolsep{\fill}}lcc
%@{\hspace{0em}}l@{\hspace{3em}}c@{\hspace{3em}}c@{\hspace{0em}}
}
\toprule
$\alpha$     & JPEG-deartifacting & Raindrop removal  \\ \midrule
 % DIP~\cite{ulyanov2018deep}     &     520      &     1.2           \\
$0.01$    & 24.56 &  21.84\\
$0.05$    &      25.07 & 22.18    \\
$0.1$    &      24.38 & 21.64    \\

\bottomrule
\end{tabular*}
\end{table}

\noindent\textbf{Efficiency Comparison}
In Table \ref{tab:additional_comparisons_runtime}, we compare the runtime and memory consumption of training-free methods. \methodName\ achieves a good balance between image quality and computational efficiency while offering a broader range of applicability compared to most existing methods.

\begin{table}[]
\centering
\caption{Runtime (in seconds) and Memory consumption (in Gigabytes) comparison of training-free methods on CelebA. The input image is of size $256 \times 256$. \label{tab:additional_comparisons_runtime}}
     \centering
    \scriptsize
    \captionsetup{font=scriptsize}
%\vspace{-0.3cm}
\begin{tabular*}{\linewidth}{
@{\extracolsep{\fill}}lcc
%@{\hspace{0em}}l@{\hspace{6em}}c@{\hspace{8em}}c@{\hspace{0em}}
}
\toprule
Method     & Runtime [s] & Memory [GB]   \\ \midrule
 % DIP~\cite{ulyanov2018deep}     &     520      &     1.2           \\
 GDP~\cite{fei2023generative}     &       168  &       1.1              \\
BIRD~\cite{chihaoui2024blind}     &    234  &    1.2   \\
BlindDPS~\cite{chung2023parallel} & 270 & 6.1\\
DreamClean~\cite{xiaodreamclean} & 125 & 1.3\\

\methodName    &      190 &      1.2         \\

\bottomrule
\end{tabular*}
\end{table}

%\vspace{-0.4cm}
\section{Conclusion}
We introduced \methodName, a novel zero-shot, training-free image restoration method that performs effectively even in fully blind settings. By leveraging a pre-trained diffusion model as a robust image prior, \methodName restores inverted input images through a novel approach that guides noise samples toward higher-density regions of the standard normal distribution. \methodName achieves state-of-the-art performance across various blind restoration tasks and demonstrates broad applicability, making it a promising solution for real-world image restoration.

% We have introduced \methodName a novel zero-shot image retoration method the degradation operator is unknown. By leveraging a pre-trained diffusion model as a robust image prior and introducing a novel approach to guiding the noise sample towards higher-density regions, \methodName successfully restores images with high fidelity across various degradation types. Our experiments demonstrate that \methodName outperforms existing fully blind methods in different blind restoration tasks. \methodName offers a generalizable and training-free solution, making it a promising approach for practical image restoration in diverse, real-world settings.

% end todo

{
    \small
    \bibliographystyle{ieeenat_fullname}
    \bibliography{main}
}

\clearpage
\appendix

\section{Proof of Proposition 1 \label{props0}}

% \begin{proposition}
% Given a deterministic mapping $G_\text{DDIM}$ induced by a pre-trained diffusion model, we solve the following optimization problem
% \begin{align}\label{eq:constrained_z_3}
%     \tilde z = \arg \min_{z \in \mathbb{R}^{n \times m \times c}} \| y - G_\text{DDIM}(z) \|^2.
% \end{align}
% When $y$ is a degraded image, the solution $ \tilde z$ of Problem~\eqref{eq:constrained_z_3} lies in the low-density region $\mathcal{R}_L$ of $\mathcal{N}(0, \mathbf{I})$.
% \end{proposition}
% \begin{proposition}
% Let $y$ be a degraded image and let us solve the following optimization problem
% \begin{align}\label{eq:constrained_z_3}
%     \tilde z = \arg \min_{z \in \mathbb{R}^{n \times m \times c}} \| y - G_\text{DDIM}(z) \|^2.
% \end{align}
% Since $G_\text{DDIM}$ is invertible and differentiable, its determinant of the Jacobian admits an upper bound $B$. Then, by setting $\tau_\alpha= \tau B$, the solution $ \tilde z$ of Problem~\eqref{eq:constrained_z_3} lies in a low-density region $\mathcal{R}_L$ of $\mathcal{N}(0, \mathbf{I})$.
% \end{proposition}
%\begin{proposition}
\noindent\textbf{Proposition 1.} 
\textit{Let $y$ be a degraded image and let us define 
\begin{align}
    \tilde z = G_\text{DDIM}^{-1}(y).\label{eq:constrained_z_3}
\end{align}
%solve the following optimization problem
% \begin{align}\label{eq:constrained_z_3}
%     \tilde z = \arg \min_{z \in \mathbb{R}^{n \times m \times c}} \| y - G_\text{DDIM}(z) \|^2.
% \end{align}
Since $G_\text{DDIM}$ is invertible and differentiable, the determinant of its Jacobian admits an upper bound $B$. Then, by setting $\tau_\alpha= \tau B$, the noise $ \tilde z$ in \cref{eq:constrained_z_3} lies in a low-density region $\mathcal{R}_L$ of $\mathcal{N}(0, \mathbf{I})$.}\\
%\end{proposition}
\textit{Proof.}
Because the image $y$ is degraded, we have that \begin{align}
p_x(y)<\tau.\label{eq:degraded}    
\end{align} 
Also, since the mapping $g(z) = G_\text{DDIM}(z)$ is invertible and differentiable, we have that 
$$p_z(z) = p_x(g(z)) |\nabla g(z)|.$$
We also have that $y = g(\tilde z)$ from  \cref{eq:constrained_z_3}.
Recall that the determinant of the Jacobian of $g$, $|\nabla g(\tilde z)|$,  admits an upper bound $B$. Then, we have that
\begin{align}
    p_z(\tilde z) = p_x(g(\tilde z)) |\nabla g(\tilde z)|<p_x(y) B<\tau B,
    \label{eq:bound}
\end{align}
where the last inequality used \cref{eq:degraded}.
If we set $\tau_\alpha = \tau B$, then \cref{eq:bound} implies that $\tilde z\in \mathcal{R}_L$.
%The pre-trained model is trained to map standard normal noise (from a high-density region of $\mathcal{N}(0, \mathbf{I})$) to \textbf{clean} images. Suppose $\tilde z$ lies in a high-density region of a standard normal $\mathcal{N}(0, \mathbf{I})$, and, by assuming Problem~\eqref{eq:constrained_z_3} is solved optimally (because of the invertibility of $G_\text{DDIM}$, and the domain of $z$, Problem~\eqref{eq:constrained_z_3} can always be optimized to reach the absolute minimum), $ G_\text{DDIM}(\tilde z) = y$, which implies that $y$ is by definition a clean image. However, this leads to a contradiction. Thus $\tilde z$ lies in a low-density region of a standard normal $\mathcal{N}(0, \mathbf{I})$

\section{An Efficient Computation of $G_\text{DDIM}(z)$ \label{efficient-compute}}

 While we adopt a similar approach to \cite{chihaoui2024blind}, For efficient diffusion inversion, we use $\delta t = 100$, which defines the step increments in the diffusion process.

\begin{algorithm}[H]
 \algsetup{linenosize=\small}
  \scriptsize
  \caption{ $G_\text{DDIM}( z_T, \delta t)$\label{alg:ddim}}
  \begin{algorithmic}[1]
  \STATE $t = T$ 
  \WHILE{$t > 0$}  
    \STATE $\hat{x}_{0|t} = (z_t - \sqrt{1 - \bar{\alpha}_{t}}  \epsilon_{\theta} (z_t, t))/\sqrt{\bar{\alpha}_{t}}$ 
 
  \STATE $z_{t- {\delta t}} = \sqrt{\bar{\alpha}_{t-{\delta t}}}  \hat{x}_{0|t} + \sqrt{1 - \bar{\alpha}_{t-{\delta t}}} . \frac{z_t - \sqrt{\bar{\alpha}_{t}} \hat{x}_{0|t}}{\sqrt{1 - \bar{\alpha}_{t}}}$
 
   \STATE $t \leftarrow t - \delta t$
  \ENDWHILE\\
 \STATE return $\hat x_{0}$
\end{algorithmic}
\end{algorithm}

\section{Review on Statistical Testing \label{statistical-testing}}

\subsection{General Review}

Statistical testing is a method used to infer properties of a population based on sample data. It begins by formulating two competing hypotheses: \textit{the null hypothesis} (\(H_0\)) and the \textit{alternative hypothesis} (\(H_1\)). The null hypothesis represents the default assumption, such as no effect or no difference, while the alternative suggests a deviation from this assumption.
The process involves computing a test statistic, a value derived from the data that quantifies the evidence against \(H_0\). The p-value, representing the probability of obtaining a result as extreme as observed under \(H_0\), is compared against a significance level (\(\alpha\)). 

\subsection{Normality Test}

In our case, we use the D’Agostino and Pearson normality test~\cite{diagostino1971omnibus}, which is a statistical method used to assess whether a set of measurements follows a normal distribution. It combines skewness and kurtosis to evaluate the departure of a distribution from normality. 

The test works as follows:

1. \textit{Hypotheses}:
    Null hypothesis (\(H_0\)): The data follows a normal distribution.
   Alternative hypothesis (\(H_1\)): The data does not follow a normal distribution.

2. \textit{Calculate Test Statistics}:
   Compute the \textit{skewness} (\(g_1\)) and \textit{kurtosis} (\(g_2\)) of the set of measurements \{$x_1, .., x_n$\} where $\bar{x}$ is its mean:
     \[
     g_1 = \frac{\frac{1}{n} \sum_{i=1}^n (x_i - \bar{x})^3}{\left(\frac{1}{n} \sum_{i=1}^n (x_i - \bar{x})^2\right)^{3/2}}
     \]
     \[
     g_2 = \frac{\frac{1}{n} \sum_{i=1}^n (x_i - \bar{x})^4}{\left(\frac{1}{n} \sum_{i=1}^n (x_i - \bar{x})^2\right)^2} - 3
     \]

3. \textit{Transform to Normalized Variables}:
   - Skewness is transformed to \(z_1\), and kurtosis to \(z_2\), using transformations that normalize these statistics for small sample sizes. These steps adjust for the bias in skewness and kurtosis estimates.

4. \textit{Combine into the Omnibus Test Statistic}:
   The test statistic \(K^2\) combines both \(z_1^2\) and \(z_2^2\) into
     \[
     K^2 = z_1^2 + z_2^2.
     \]
   This statistic follows a \(\chi^2\)-distribution with 2 degrees of freedom under the null hypothesis.

5. \textit{Decision Rule}:

   - Compute the p-value:
     \[
     \text{p-value} = P(\chi^2 > K^2 \mid H_0)
     \]
     
   - Reject \(H_0\) if \(\text{p-value} < \alpha\).

\noindent{To conduct the normality test, we used the built-in implementation of Numpy \cite{harris2020array} and we set $\alpha=0.05$.}

\end{document}